\pdfoutput=1

\documentclass[11pt,hyphens]{article}

\usepackage[final]{acl}

\usepackage{times}
\usepackage{latexsym}
\usepackage{spverbatim}
\usepackage{xcolor}
\usepackage{colortbl}

\usepackage[T1]{fontenc}

\usepackage[utf8]{inputenc}

\usepackage{microtype}

\usepackage{inconsolata}

\usepackage{graphicx}

\usepackage{bbm}
\usepackage{subcaption}
\usepackage{multirow}
\usepackage{enumerate}
\usepackage{amsmath}
\usepackage{microtype}
\usepackage{amsmath}
\usepackage{amsthm,amssymb}
\usepackage{graphicx}
\usepackage{booktabs} 
\usepackage{multirow}
\usepackage{array}
\usepackage{arydshln}
\usepackage{color}
\usepackage{xcolor}
\usepackage{url}

\usepackage{hyperref}

\author{
 \textbf{Mengyu Wang\textsuperscript{1}} \quad
 \textbf{Sotirios Sabanis\textsuperscript{1,2,3}} \quad
 \textbf{Miguel de Carvalho\textsuperscript{1,4}} \\
 \textbf{Shay B. Cohen\textsuperscript{1}} \quad
 \textbf{Tiejun Ma\textsuperscript{1}}
\\
\textsuperscript{1}The University of Edinburgh, United Kingdom\\
\textsuperscript{2}National Technical University of Athens, Greece\\
\textsuperscript{3}Archimedes/Athena Research Centre, Greece\\
\textsuperscript{4}University of Aveiro, Portugal\\
 \medskip
 \texttt{\{mengyu.wang, s.sabanis, miguel.decarvalho, scohen, tiejun.ma\}@ed.ac.uk}
}

\newenvironment{enumeratesquish}[2]{
  \begin{list}{\labelenumi}{
    \setlength{\itemsep}{#1}
    \setlength{\labelwidth}{#2}
    \setlength{\leftmargin}{\labelwidth}
    \addtolength{\leftmargin}{\labelsep}
    \setlength{\parsep}{0em}
    \setlength{\topsep}{0em}
  }
}{\end{list}}

\title{One More Question is Enough, Expert Question Decomposition (EQD) Model for Domain Quantitative Reasoning}

\begin{document}
\maketitle
\begin{abstract}
Domain-specific quantitative reasoning remains a major challenge for large language models (LLMs), especially in fields requiring expert knowledge and complex question answering (QA). 
In this work, we propose Expert Question Decomposition (EQD), an approach designed to balance the use of domain knowledge with computational efficiency. EQD is built on a two-step fine-tuning framework and guided by a reward function that measures the effectiveness of generated sub-questions in improving QA outcomes.
It requires only a few thousand training examples and a single A100 GPU for fine-tuning, with inference time comparable to zero-shot prompting. Beyond its efficiency, EQD outperforms state-of-the-art domain-tuned models and advanced prompting strategies.
We evaluate EQD in the financial domain, characterized by specialized knowledge and complex quantitative reasoning, across four benchmark datasets. Our method consistently improves QA performance by 0.6\% to 10.5\% across different LLMs. Our analysis reveals an important insight: in domain-specific QA, a single supporting question often provides greater benefit than detailed guidance steps.
\end{abstract}

\section{Introduction}
The performance of LLMs may significantly degrade in specialized domains~\cite{shen2024tag}. Even advanced LLMs, such as GPT-4o~\cite{hurst2024gpt} and Llama3~\cite{dubey2024llama}, exhibit substantial gaps compared to human experts in domain-specific question answering (QA), particularly in tasks involving quantitative reasoning, like financial analysis~\cite{chen2021finqa}. This performance gap stems from the complex terminology and specialized knowledge inherent in these domains, which are often underrepresented in the pretraining corpora used for general-purpose LLMs.

\begin{figure}[t]
\centering
\begin{center}
   \includegraphics[width=0.92\linewidth]{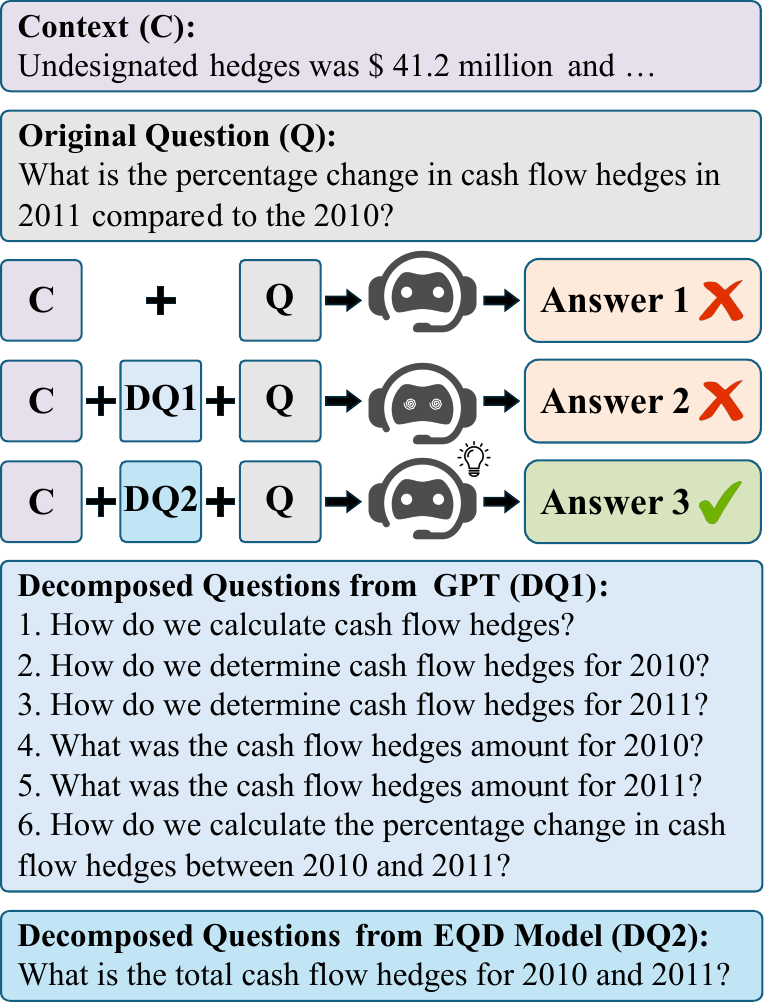}
    \vspace{-0.0cm}
   \captionof{figure}{A practical example comparing different QA processes. General LLMs struggle to give correct answers directly. The CoT method attempts to simplify the question, but often decompose the query into overly detailed steps, introducing confusion. In contrast, our EQD model adds a single sub-question that effectively guides the LLM toward the correct answer.}
\label{fig:overview}
\end{center}
\vspace{-0.6cm}
\end{figure}

Recent research addresses domain quantitative reasoning challenges through two main approaches: domain-adapted fine-tuning~\cite{wang2023fingpt, yang2023investlm} and prompting techniques~\cite{wei2022chain, chen2022program}. However, both approaches face significant limitations. Domain fine-tuning is resource-intensive, requiring large, high-quality domain datasets and significant computational power~\cite{wu2023bloomberggpt}. Moreover, many state-of-the-art models like GPT are closed-source, making it difficult to tailor the model to the domain. Prompt-based methods, while model-agnostic and training-free, often reduce inference efficiency due to long augmented inputs. Moreover, they are constrained by the limited extra knowledge contained only in the prompt itself~\cite{srivastava2024evaluating}.

Two often-overlooked aspects are potential to mitigate these limitations. First, complex domain knowledge can often be decomposed into simpler, more general components. For instance, a financial question like ``\emph{what is the ROI of the investment}'' can be transformed into basic arithmetic questions about ``\emph{the initial investment, returns, and percentage change}''. This suggests that much of the domain-specific knowledge encoded in fine-tuned models may be redundant. Second, given LLMs' inherent strong reasoning abilities, detailed, step-by-step guidance may be unnecessary, or even detrimental. Overly verbose reasoning chains can introduce noise or distract from the core problem, suggesting that targeted questions focusing on key challenges tend to be more effective.

Based on these insights, we develop an Expert Question Decomposition (EQD) model that generates concise and effective supporting questions to guide LLMs in domain-specific reasoning tasks. We select the financial domain, which is characterized by specialized knowledge and quantitative requirements, as the testbed for domain quantitative reasoning. As illustrated in Figure~\ref{fig:overview}, a challenging financial question that remains unsolvable using standard decompositions becomes solvable with a single, supporting question from our EQD model. We consistently observe that such questions significantly improve QA performance.

EQD is developed through a two-step process: \emph{domain fine-tuning} and \emph{QA expert alignment}. In the first step, we fine-tune Llama 3.1-8B-Instruct model using step-by-step question data from financial dialogues. Unlike prior approaches that aim to inject broad domain knowledge into LLMs, our method focuses specifically on fine-tuning the model to decompose domain questions into simpler sub-questions. In the second step, we use a reward-based alignment process. We design a novel reward function that measures the impact of supporting questions by comparing QA performance with and without decomposition. This reward guides the model through reinforcement learning to optimize the quality of generated sub-questions.

Our approach balances computational efficiency and domain knowledge integration. It requires only a small decomposition dataset and a representative domain QA dataset, substantially less than domain-specific LLM fine-tuning. EQD is also compatible with both open- and closed-source LLMs, unlike domain-specific models such as FinMA~\cite{xie2023pixiu} and InvestLM~\cite{yang2023investlm}, which are tied to outdated base models. During inference, EQD typically add only a single supporting question, incurring minimal additional processing overhead and preserving response time comparable to zero-shot prompting.

Beyond its efficiency, EQD demonstrates strong performance in improving domain QA. Across four financial QA benchmarks, it achieves performance gains of 0.6\% to 10.5\% across multiple LLMs, outperforming advanced domain-adapted models and prompting techniques. These results challenge the conventional emphasis on comprehensive and step-by-step CoT prompting, revealing that concise and supporting questions can lead to better LLM reasoning in specialized domains.

In summary, our key contributions include:
\begin{enumeratesquish}{0em}{0.5em}
    \item[1] We propose a two-stage training framework for expert question decomposition, which integrates domain knowledge efficiently while maintaining inference time comparable to zero-shot prompting. It requires only a few thousand examples and one GPU for training. 
    
    We have made our code publicly available at: \href{https://github.com/astudentuser/Expert-Question-Decomposition-EQD-Model-for-Domain-Quantitative-Reasoning}{EQD GitHub repository}.
    \item[2] We introduce a novel answer comparison reward that guides EQD to generate concise and effective supporting questions.
    Experiments on four financial QA datasets show that our method consistently improves the performance of various LLMs by 0.6\% to 10.5\%, and achieve at least a 5\% improvement over existing QA approaches.
    \item[3] Our results and analysis reveal that concise and supporting questions are more effective than extensive reasoning steps, providing new insights into LLM reasoning mechanisms.
\end{enumeratesquish}

\section{Related Work}
\label{sec:related_work}

\begin{figure*}[t]
\centering
\begin{center}
   \includegraphics[width=0.98\linewidth]{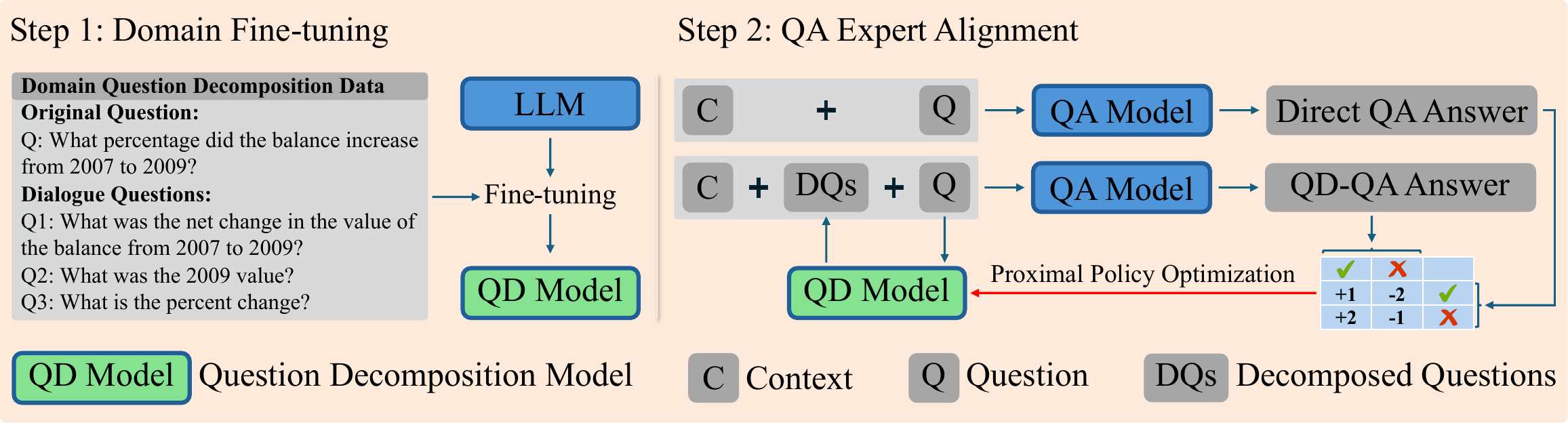}
   \vspace{-0.1cm}
   \captionof{figure}{Two-step training framework of the Expert Question Decomposition model.}
\label{fig:method}
\end{center}
\vspace{-0.7cm}
\end{figure*}

LLMs have shown  progress in quantitative reasoning~\cite{achiam2023gpt, zelikman2022star}, but continue to face challenges in incorporating specialized domain knowledge~\cite{shen2024tag}. The financial domain, with its combination of technical terminology and numerical reasoning, has emerged as a key testbed for evaluating domain-specific reasoning capabilities. NLP techniques play an important role in advancing financial applications~\cite{wang2024mana, wang2024modeling}, and several financial benchmark datasets have been introduced to evaluate different aspects of domain quantitative reasoning. FinQA~\cite{chen2021finqa} and ConvFinQA~\cite{chen2022convfinqa} focus on multi-step mathematical reasoning, while TATQA~\cite{zhu2021tat} addresses the challenge of processing diverse input formats, including structured tables and unstructured text in question answering. LLMs have prompted the development of various methods to enhance domain reasoning, including specialized QA models~\cite{zhao2022multihiertt, herzig2020tapas}, prompting techniques~\cite{singh2024finqapt,leang2024comat}, and reasoning planning frameworks~\cite{srivastava2024evaluating}.

However, most existing methods primarily depend on manually designed prompts to optimize LLMs’ domain performance~\cite{li2024mtmt, cao2023step, huang2023question, li2025fingearfinancialmappingguidedenhanced}, without fully considering the differences between human reasoning and model behavior. Some recent studies have explored using LLMs’ own errors as optimization signals~\cite{wu2025composerag}, but their reward functions remain relatively simple. In this work, we propose a four-level reward function that guides LLMs to improve the generation of supporting questions by leveraging QA model outputs. This design aligns with QA models’ inherent challenges, thereby enhancing LLM performance.

Despite these advances, a persistent challenge across these approaches lies in balancing domain knowledge integration with computational efficiency. Our EQD method addresses this trade-off by offering a lightweight yet effective question decomposition model. Rather than relying on extensive domain fine-tuning or verbose prompting, EQD generates concise, supporting questions that enhance LLM reasoning with minimal computational overhead.

\section{Method}
\label{sec:method}
We propose a two-step training approach for developing the Expert Question Decomposition (EQD) model, as illustrated in Figure~\ref{fig:method}. The first step involves instruction fine-tuning to integrate domain knowledge into the model. Unlike existing methods that require extensive domain corpora for LLM adaptation, we propose to use a small financial conversation dataset to develop a domain-specific Question Decomposition (QD) model. The second step uses reinforcement learning with Proximal Policy Optimization (PPO; \citealt{schulman2017proximal}) to align the QD model with the QA process, optimizing the effectiveness of generated Decomposed Questions (DQs) in enhancing QA performance. {We introduce a novel answer comparison reward function for this end.

\subsection{Domain Fine-tuning}
Domain knowledge plays an important role in improving LLM performance on specialized tasks. Traditional methods incorporate domain knowledge by fine-tuning on extensive domain-specific corpora or linking external databases. However, the gap between general and domain knowledge is often narrower than expected. In many cases, a simple term explanation can transform domain-specific questions into general queries. For instance, a specialized financial question like ``\emph{What is the ROI of the investment}'' can be converted to a general arithmetic question involving ``\emph{initial investment, returns, and percentage change}''.

Motivated by this insight, we propose to fine-tune a lightweight LLM (Llama3.1-8B-Instruct in our experiments) exclusively for domain-specific question decomposition. This approach balances domain adaptation and training efficiency.

We use ConvFinQA~\cite{chen2022convfinqa}, an expert-annotated financial conversation dataset containing around only 3,000 entries, to fine-tune our QD model. As shown on the left side of Figure~\ref{fig:method}, we extract only the original questions and the dialogue questions as input information, discarding answers and explanations. We design a system prompt focused solely on question decomposition (detailed in Appendix~\ref{sec:appendix_prompt}), and fine-tune the base LLM using next-token prediction on this input.

The resulting QD model embeds domain-specific knowledge essential for understanding and decomposing financial questions. This mitigates the limitations of prompt-based methods, which are constrained by context length and limited domain-specific input. The QD model serves as an automatic reasoning chain generator, generating effective supporting questions to assist QA process. 

Since the model is trained only for question decomposition, its training cost is significantly lower than full-domain LLM fine-tuning. Additionally, although developed based on an open-source LLM, this QD model can transfer its domain knowledge to any other LLMs, regardless of their open- or closed-source nature. By inserting its generated decomposed questions into QA inputs, we observe consistent improvements in reasoning accuracy across diverse LLMs.

\subsection{QA Expert Alignment}
\label{sec:QA_alignment}
While domain fine-tuning equips the QD model with financial question decomposition expertise, it does not guarantee optimal support for the QA process. An ideal QD model should generate supporting questions that increase the QA model's likelihood of answering correctly while minimizing the risk of introducing misleading information. This requirement parallels the broader LLM alignment problem, where the goal is to align a model’s behavior with human intent. Similarly, we aim to align the QD model’s outputs with the latent preferences of the QA model, rather than solely mimicking human-annotated decomposition patterns.

To achieve this QA alignment, we introduce an answer comparison reward to quantify the impact of the QD model's outputs on QA performance. Specifically, we compare two outputs from the QA model: one answer obtained through direct QA (i.e., the LLM answers the question without assistance), and another answer from QD-assisted QA (i.e., the same LLM answers with supporting questions generated by the QD model). This controlled setup, shown on the right side of Figure~\ref{fig:method}, ensures that the only difference lies in the presence of decomposed questions (DQs), isolating their influence.

Let $a_{di}$ denote the direct QA answer and $a_{qd}$ represent the QD-assisted answer. The reward score $r$ is calculated as:
\begin{align}
\label{eq:reward}
&c(a) = 
\begin{cases}
    1,& \text{if answer $a$ is correct} \\
    -1,              & \text{otherwise}
\end{cases}
,\\
& r = c(a_{qd}) \cdot (1 + 0.5 \cdot \lvert c(a_{di}) - c(a_{qd}) \rvert)
\end{align}
where $c(\cdot)$ is the function to evaluate the correctness of the given answer.

This reward yields four possible values: +2, +1, -2, and -1, corresponding to high positive, low positive, high negative, and low negative rewards, respectively. DQs receive a positive score when they lead to correct answers, and a negative score otherwise. The magnitude of the score (high or low) is determined by whether the DQs alter the correctness of the answer compared to direct QA. Specifically, the four values represent: +2: DQs correct an originally incorrect answer, +1: DQs preserve a correct answer, -2: DQs turn a correct answer into an incorrect one, -1: DQs preserve an incorrect answer. This scoring mechanism captures both the correctness and influence of the DQs, encouraging outputs that improve QA performance and penalizing those that degrade it.

Using this reward, we fine-tune the QD model via PPO algorithm. PPO adjusts model parameters to maximize the expected reward while maintaining a bounded KL divergence from a reference model, ensuring updates remain within a trust region. The reference model is initialized as a copy of the QD model from step 1, allowing us to preserve its financial knowledge and decomposition style.

In summary, this reinforcement learning stage aligns the QD model with the QA model’s requirements, evolving it into the EQD model to generate decomposed questions that effectively support the QA model in producing correct answers.

\subsection{Resource Requirement}
\label{sec:peft}
We detail the resource requirements of our method to highlight its efficiency and practicality.

\noindent\textbf{Training Data.} Our approach requires only a question decomposition dataset and a representative QA dataset for the focused domain. After completing the two-step training, the resulting EQD model can generalize to various QA datasets within the same domain. In our experiments, we use just two datasets, ConvFinQA and FinQA, among the many datasets used for financial LLM fine-tuning~\cite{xie2024open}, yet demonstrate strong performance across four different financial QA benchmarks.

\noindent\textbf{Computational Resources.} Our training requires only a single GPU capable of fine-tuning the base model (an A100 in our setup). Although the two-step training involves multiple roles-the QA model, QD model, and reference model-we efficiently organize model parameters to keep the resource demands equivalent to running a single LLM.

We use Low-Rank Adaptation (LoRA; \citealt{hu2022lora}) for parameter-efficient fine-tuning in both steps. The added adapter contains only 22 million parameters, about 0.27\% of the 8B base model. We use a continuous fine-tuning strategy, training the same adapter throughout both steps. In step 1, the model comprises the base LLM with a trainable adapter. In step 2, although three logical models are involved, only one full LLM and two adapters are required in practice. Specifically, the base LLM serves as the QA model, the trainable adapter forms the QD model, and a frozen copy of the adapter serves as the reference model. These roles are managed by activating or deactivating the corresponding adapters, minimizing memory overhead.

In summary, our method integrates domain knowledge in a more resource-efficient manner than existing domain model fine-tuning methods.

\subsection{Expert Question Decomposition Model}
After the two-step fine-tuning, we develop an Expert Question Decomposition (EQD) model combining domain expertise with optimized QA alignment. This model offers two key advantages over existing prompting-based methods.

First, the EQD model generates prompts automatically, in contrast to conventional methods that rely on rule-based or manually crafted prompts. This automation enables broad applicability across diverse datasets within the same domain, eliminating the need for dataset-specific prompt design.

Second, while existing prompting methods often construct detailed and comprehensive guidance, our EQD model produces concise yet effective supporting questions. As illustrated in Figure~\ref{fig:overview}, a single well-chosen supporting question outperforms multiple detailed guidance steps. This observation suggests that LLMs already possess strong reasoning capabilities, and excessive guidance can be redundant, or even harmful. Our training objectives do not explicitly penalize or limit generation length. Instead, the reinforcement learning reward function solely optimizes the effectiveness of the generated decomposed questions in improving QA outcomes. The natural emergence of conciseness in the model’s outputs indicates that brevity is an inherent requirement of effective QA support.

In conclusion, our EQD model exhibits two key advantages over existing prompting methods: domain-specific versatility and concise yet effective question decomposition. 

\section{Experiment Settings}
\label{sec:experiment}
\subsection{Training and Testing Datasets}
\label{sec:datasets}
Our EQD model is trained using two financial datasets. For step 1, we use the training split of ConvFinQA~\cite{chen2022convfinqa}, comprising 3,073 entries of financial reasoning conversations. For step 2, we use the training split of FinQA~\cite{chen2021finqa}, containing 6,250 financial QA pairs.

To evaluate the generalized QA improvement capability of our EQD model, we conduct testing on four distinct financial datasets: FinQA, TAT-QA~\cite{zhu2021tat}, ECTQA~\cite{mukherjee2022ectsum}, and EDTQA~\cite{xie2024finnlp}. All four testing datasets require both domain-specific knowledge and numerical reasoning capabilities. A detailed description of each dataset, along with relevant statistics, is provided in Appendix~\ref{sec:appendix_dataset}.

\subsection{Implementation Details}
\label{sec:implementation}
We use the Llama3.1-8B-Instruct model~\cite{dubey2024llama} as both the base model and QA model. For reinforcement learning, we use the PPO model’s value head as the critic model. For performance evaluation, we use the exact match accuracy (EmAcc) metric, following established practices in previous works~\cite{xie2023pixiu, zhao2024revolutionizing}.  Since the key points of the answers are numerical values, we implement a systematic evaluation process~\cite{singh2024finqapt} that extracts values and compares them to the ground truth. Detailed information on parameter settings, API costs, evaluation setup, and computing devices are presented in Appendix~\ref{sec:appendix_implementation}. The strategy for managing trainable parameters across the two fine-tuning stages is discussed in Appendix~\ref{sec:appendix_ft_strategy}.

In step 1, we perform next-token prediction by concatenating a question decomposition instruction, the original question, and the conversation sub-questions as inputs. In step 2, we conduct QA by concatenating a financial QA instruction, the financial article, the decomposed questions, and the final questions. The model's final response serves as the answer to the original question. Detailed prompt examples are provided in Appendix~\ref{sec:appendix_prompt}.

In Step 2, we assign specific scores of +2, +1, –1, and –2 to the four reward levels. We also experimented with alternative discrete reward configurations for comparison-based learning, such as (+2, +1, –1, –4) and (+4, +1, –1, –2). The results show that the balanced configuration (+2, +1, –1, –2) achieves the best performance. Other arrangements, such as merging the lower levels or using unbalanced scores, led to performance degradation. Consequently, we use the balanced reward configuration for our final method.

To evaluate the generalization ability of our approach for obtaining an EQD model, we also experiment with three additional LLMs: Llama3.2-1B-Instruct, Llama3.2-3B-Instruct, and DeepSeek-R1-Distill-Qwen-7B~\cite{deepseekai2025deepseekr1incentivizingreasoningcapability}, spanning different model sizes and architectures. The corresponding results and analysis are presented in Appendix~\ref{sec:appendix_other_base}.

\begin{table*}[t]
\vspace{-0.0cm}
\begin{center}
\begin{tabular}[t]{l r@{\hspace{8pt}} c@{\hspace{12pt}} r@{\hspace{8pt}} r@{\hspace{12pt}} r@{\hspace{8pt}} r@{\hspace{12pt}} r@{\hspace{8pt}} r@{\hspace{10pt}} r@{\hspace{8pt}} r@{\hspace{10pt}}}
\toprule
\multirow{2}{*}{Model} &
 \multicolumn{2}{c}{FinQA} & \multicolumn{2}{c}{TAT-QA} & \multicolumn{2}{c}{ECTQA} & \multicolumn{2}{c}{EDTQA} & \multicolumn{2}{c}{\textit{Average}} \\
  & Direct & EQD & Direct & EQD & Direct & EQD & Direct & EQD & Direct & EQD \\
\cmidrule(r){1-11}
Llama3.1-8B      & 47.2 & \underline{54.0} & 51.2 & \underline{54.9} & 61.8 & \underline{64.0} & 52.2 & \underline{55.1} & 53.1 & \underline{57.0} \\
GPT-3.5-turbo    & 28.4 & \underline{55.1} & 47.2 & \underline{52.7} & 64.7 & \underline{65.4} & 56.0 & \underline{57.3} & 47.1 & \underline{57.6} \\
GPT-4o           & 58.2 & \underline{62.4} & 59.1 & \underline{63.2} & 68.1 & \underline{72.5} & \underline{\textbf{64.9}} & 63.4 & 62.5 & \underline{65.4} \\
Claude3.5-sonnet & 72.9 & \underline{\textbf{73.7}} & 63.3 & \underline{\textbf{64.4}} & 74.8 & \underline{\textbf{75.2}} & 60.8 & \underline{61.2} & 67.9 & \underline{68.5} \\
\textit{Average} & 51.7 & \underline{61.2} & 55.2 & \underline{58.8} & 67.4 & \underline{69.3} & 58.5 & \underline{59.3} & 58.2 & \underline{62.1} \\

\hdashline

FinMA            & \underline{11.3} & 10.5 & \underline{19.1} & 18.2 &  \underline{1.9} &  1.8 & \underline{37.4} & 35.1 & \underline{17.4} & 16.4 \\
o3-mini      & \underline{70.0} & 67.6 & \underline{62.5} & 57.3 & \underline{74.4} & 70.2 & \underline{64.7} & 41.3 & \underline{67.9} & 59.1 \\

\bottomrule
\end{tabular}
\vspace{-0.2cm}
\captionof{table}{Comparison of LLM performance on financial QA tasks: Direct QA vs. QA supported by our EQD model. Underlined values indicate the higher score between Direct QA and EQD-supported QA. Bold values denote the best performance for each dataset.}
\label{tab:llm_resuls}
\end{center}
\vspace{-0.4cm}
\end{table*}

\subsection{Baseline Methods}
\label{sec:baseline}
We benchmark our method across various QA models and reasoning support techniques.

Our experiments use a diverse range of LLMs, including Llama3.1-8B-Instruct, GPT-3.5-turbo, GPT-4o, o3-mini, Claude3.5-sonnet\footnote{\url{https://www.anthropic.com/news/claude-3-5-sonnet}}, and FinMA~\cite{xie2023pixiu}. This selection encompasses advanced open-source, closed-source, and domain-specific fine-tuned models. We compare each model’s QA performance with and without support from our EQD method to demonstrate the generalized effectiveness of our method.

For reasoning support baselines, we compare EQD with several established prompting strategies, including  zero-shot Chain-of-Thought (0-CoT; \citealt{wei2022chain}), decomposed prompting (DP; \citealt{khot2022decomposed}), question decomposition CoT (QD-CoT; \citealt{zhou2022least}), retrieval CoT (R-CoT; \citealt{trivedi2022interleaving}), and few-shot in-context learning (N-shot; \citealt{li2023few}). These approaches are widely adopted for general QA tasks, and both 0-CoT and few-N-shot methods have proved effective in financial domains~\cite{srivastava2024evaluating}. Implementation details for all baseline methods are provided in Appendix~\ref{sec:appendix_baseline}.

\section{Results and Discussions}
\label{sec:results_discussions}
We evaluate our EQD method from four perspectives, presenting comprehensive experimental results to support our findings. In Section~\ref{sec:results_on_llms}, we compare the performance of various LLMs on four datasets, both in direct QA and with EQD support, demonstrating that our approach consistently enhances LLM performance on domain-specific quantitative reasoning tasks. Section~\ref{sec:results_of_methods} focuses on the two most challenging datasets, FinQA and TAT-QA, to compare different reasoning support methods, further validating the effectiveness of EQD. Section~\ref{sec:ablation_study} presents ablation studies, comparing our EQD model with other LLMs for question decomposition, and analyzing the impact of each fine-tuning step. Finally, Section~\ref{sec:efficiency_analysis} evaluates the computational efficiency of our EQD methods through inference time and generation length analysis.

\subsection{Generalized QA Improvement on LLMs}
\label{sec:results_on_llms}
Table~\ref{tab:llm_resuls} presents a comparative analysis of various LLMs on different financial QA datasets, evaluated both with and without EQD support.

Our EQD model yields consistent performance improvements across all general LLMs on different datasets, with two exceptions: FinMA and o3-mini. These models operate independently and cannot take advantage of the supporting techniques due to inherent limitations.

For general LLMs, EQD-supported QA consistently outperforms direct QA in average performance across both models and datasets. And the best results on three datasets, FinQA, TAT-QA, and ECTQA, are achieved by LLMs supported by our EQD model. Importantly, despite being trained solely on the FinQA training set and fine-tuned only using Llama3.1-8B-Instruct as the QA model, the EQD model exhibits strong generalization ability, improving QA across a range of datasets and models. This underscores the robustness of the expert question decomposition strategy.

The performance of the two standalone models, FinMA and o3-mini, reflects the limitations of systems that function in isolation rather than flaws of our approach. FinMA, an open-source LLM fine-tuned specifically for financial tasks, relies on the first-generation Llama model and is constrained by a limited input window. These factors hinder its ability to benefit from any additional contextual information. The o3-mini model, OpenAI's latest reasoning model, uses a simulated reasoning mechanism. According to OpenAI's documentation, this special model performs optimally with straightforward prompts, and prompt engineering techniques may actually impede its performance\footnote{\url{https://platform.openai.com/docs/guides/reasoning-best-practices}}. These cases, representing domain-specific and reasoning-optimized models, highlight the limitations of methods that can only work independently. In contrast, our EQD method demonstrates flexibility and compatibility with a wide range of advanced LLMs to achieve optimal results.

\begin{table*}[t]
\begin{center}
\begin{tabular}[b]{l l p{1.5cm}<{\centering}@{\hspace{-4.5pt}} p{1.5cm}<{\centering}@{\hspace{-5.0pt}} p{1.5cm}<{\centering}@{\hspace{-3.5pt}} p{1.5cm}<{\centering}@{\hspace{-3.5pt}} p{1.5cm}<{\centering}@{\hspace{-3.5pt}} p{1.5cm}<{\centering}@{\hspace{-3.5pt}} p{1.5cm}<{\centering}@{\hspace{-3.5pt}} p{1.5cm}<{\centering}@{\hspace{-3.5pt}} p{1.5cm}<{\centering}} 
\toprule
\multicolumn{1}{c}{\multirow{2}{*}{\shortstack{ QA \\ LLM}}} & \multicolumn{1}{c}{\multirow{2}{*}{\shortstack{\\ Dataset }}} & \multicolumn{9}{c}{Methods} \\
\cmidrule{3-11}
 & & Direct & 0-CoT & DP1 & DP2 & QD-CoT & R-CoT & N-shot & EQD & Manual \\
\cmidrule{1-11}
\multicolumn{1}{c}{\multirow{2}{*}{\shortstack{ Llama \\ 3.1-8B}}} & FinQA & 47.2 & \underline{52.0} & 47.6 & 50.0 & 50.0 & 48.0 & 45.1 & \textbf{54.0} & 51.5 \\
& TAT-QA & 51.2 & \textbf{57.4} & 49.7 & 51.4 & 51.8 & 49.9 & 53.9 & \underline{54.9} & 51.6 \\
\hdashline
\multicolumn{1}{c}{\multirow{2}{*}{\shortstack{ GPT-3.5 \\ -turbo}}} & FinQA & 28.4 & 16.8 & 31.2 & 51.4 & 49.6 & 50.9 & 39.1 & \textbf{55.1} & \underline{52.6} \\
& TAT-QA & 47.2 & 46.8 & 46.5 & 52.6 & 46.4 & \textbf{52.9} & 50.0 & \underline{52.7} & 52.1 \\
\hdashline
\multicolumn{1}{c}{\multirow{2}{*}{\shortstack{ GPT-4o}}} & FinQA & \underline{58.2} & 53.1 & 55.8 & 49.8 & 60.3 & 49.7 & 42.6 & \textbf{62.4} & 52.5 \\
& TAT-QA & 59.1 & 58.6 & 54.4 & 50.9 & 54.2 & 51.1 & \underline{61.0} & \textbf{63.2} & 51.8 \\
\hdashline
& \textit{Average} & 48.6 & 47.5 & 47.5 & 51.0 & \underline{52.1} & 50.4 & 48.6 & \textbf{57.1} & 52.0 \\
\bottomrule
\end{tabular}
\vspace{-0.3cm}
\captionof{table}{Comparison of different methods for conducting QA tasks across different LLMs. The names of the baseline methods are abbreviated as described in Section~\ref{sec:baseline}. Since DP is a basic question decomposition baseline, we test two versions using GPT-3.5-turbo and GPT-4o as the question decomposition models, denoted as DP1 and DP2, respectively. The final column, ``Manual'', refers to the method in which we manually design question decomposition examples based on our findings for prompts, serving as an ablation study. Bold and underline values represent the best and second-best results for each row.}
\label{tab:method_results}
\end{center}
\vspace{-0.6cm}
\end{table*}

Additionally, two trends are observed in domain quantitative reasoning. (1) Greater EQD improvements for weaker QA models. The benefit of EQD is more pronounced in weaker LLMs. When averaged across datasets, Claude 3.5 Sonnet, the strongest model, shows the smallest performance gain (+0.6\%), whereas GPT-3.5-turbo, the weakest, shows the largest (+10.5\%). This suggests that weaker LLMs struggle more with complex reasoning and thus benefit more from EQD’s decomposition. (2) Greater EQD improvements for more complex reasoning tasks. Averaged across LLMs, the smallest improvement is observed on EDTQA (+0.8\%), while the largest is on FinQA (+9.5\%). Since ECTQA and EDTQA are derived from summarization datasets, they contain questions that typically require simpler reasoning or direct value extraction. In contrast, FinQA questions demand multi-step calculations (typically 3–4 operations), which LLMs often fail to handle reliably without support. This indicates that EQD provides more value in tasks involving quantitative reasoning.

These findings reinforce our claim: EQD's question decomposition enhances LLMs' QA performance by degrading reasoning challenges. In contrast, when LLMs can already manage complex tasks, additional guidance yields diminishing returns. This also explains why concise supporting questions generated by EQD often outperform more detailed or verbose prompt instructions.

\subsection{Comparison with Different Methods}
\label{sec:results_of_methods}

Table~\ref{tab:method_results} presents the comparative results of various LLM-based methods for financial QA. Due to resource constraints, we evaluated three representative QA models: Llama3.1-8B-Instruct, GPT-3.5-turbo, and GPT-4o, using two of the most challenging datasets, FinQA and TAT-QA.

Our EQD-QA method demonstrates robust performance, outperforming all other methods by an average margin of at least 5\%. It achieves the best results in four out of six specific scenarios and ranks second in the remaining case.

The first three baseline methods, 0-CoT, DP1, and DP2, aim to elicit reasoning capabilities from LLMs without incorporating external knowledge. While effective in general QA settings, they struggle with financial tasks due to insufficient domain-specific understanding. The next set of baselines, QD-CoT, R-CoT and N-shot, introduce domain knowledge through either example-based decompositions or domain-specific reasoning chains. Although these approaches improve performance in some scenarios, they fail to deliver consistent benefits across models and datasets. This inconsistency underscores a key insight: simply embedding domain knowledge into prompts is insufficient for reliable performance gains. In contrast, our EQD model is explicitly trained to optimize the effectiveness of the additional information in supporting QA process. This targeted optimization explains its superior and consistent performance over other prompt-based strategies.

\subsection{Ablation Study}
\label{sec:ablation_study}
We conduct two sets of ablation studies to assess the components and training strategy of EQD.

First, we compare EQD with alternative methods for question decomposition. As shown in Table~\ref{tab:method_results}, the columns ``DP1'', ``DP2'', and ``Manual'' represent QA results using supporting questions from different methods. DP1 and DP2 serve as both baselines and ablations, since they use general-purpose LLMs to generate decompositions. Results show that our EQD model surpasses even the recent GPT models in generating effective sub-questions. 

The ``Manual'' approach assumes foreknowledge of EQD’s key conclusion that concise and  supporting questions are more effective. We manually write five concise examples and apply them in a 5-shot prompting setup to guide LLMs' decomposition. This method outperforms many baselines, validating our finding. However, it still underperforms our EQD model. This is because our model is specifically trained to generate the most essential sub-questions for the QA process, whereas human annotators may not always be able to identify the most important reasoning steps for LLMs.

\begin{figure}[b]
\vspace{-0.6cm}
\centering
\begin{center}
   \includegraphics[width=1.0\linewidth]{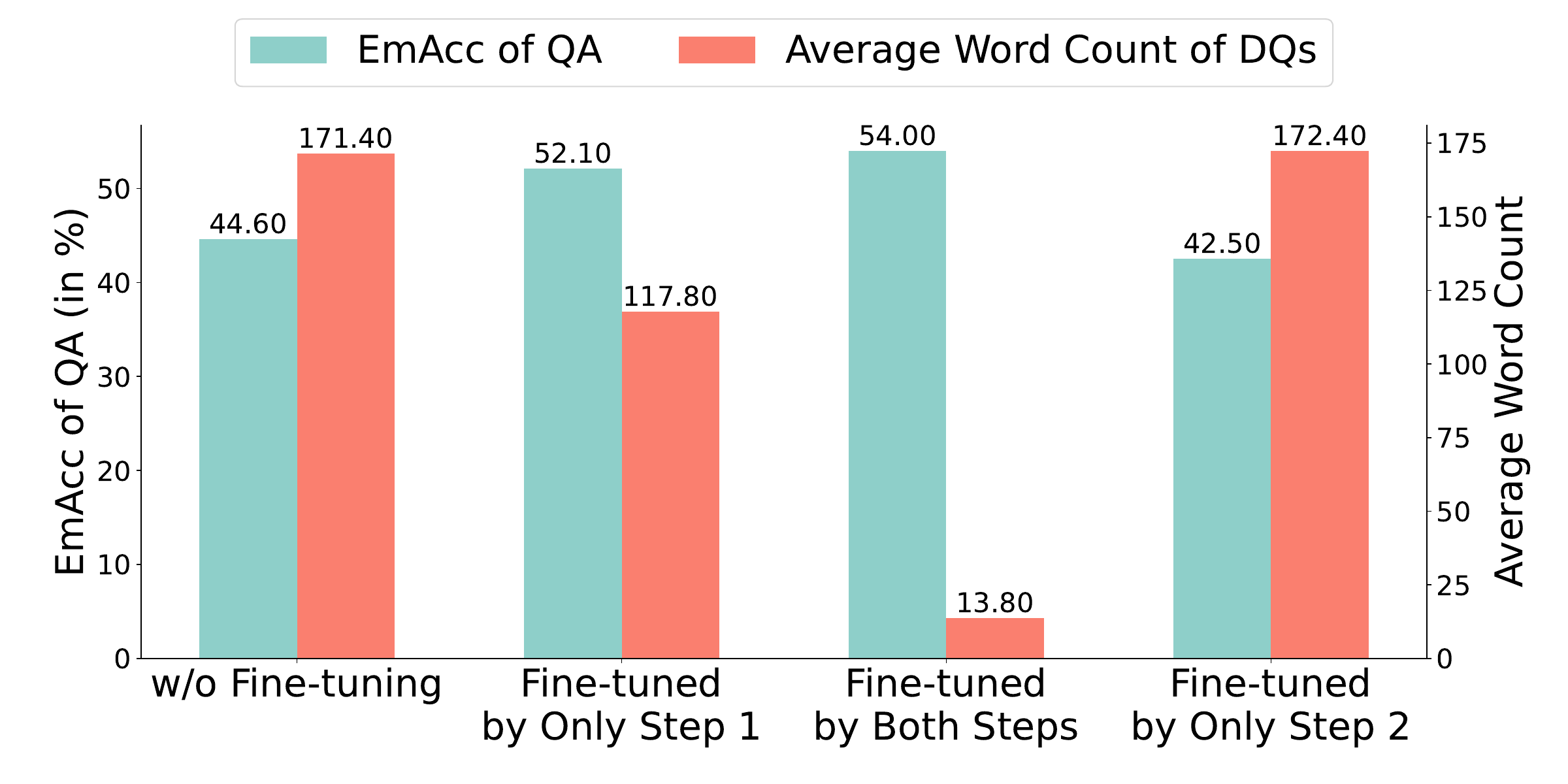}
    \vspace{-0.8cm}
   \captionof{figure}{Comparison of QD models fine-tuned differently, using Llama3.1 as QA model. Blue bars reflect QA accuracy (left y-axis), while red bars indicate the average word count of generated questions (right y-axis).}
\label{fig:ablation_steps}
\end{center}
\vspace{-0.3cm}
\end{figure}

Second, using Llama3.1-8B-Instruct as the QA model, we examine the contribution of each training step. We compare four configurations: no fine-tuning, step 1 only, both steps, and step 2 only. Figure~\ref{fig:ablation_steps} presents the QA performance and average word count of the generated sub-questions. The trend in average word count also reflects changes in the number of decomposed questions generated, with average question counts of 15.0, 6.23, 1.2, and 15.6 for the four settings, respectively.

Results indicate that combining both steps gives the best performance and most concise questions. Both steps enhance the effectiveness and conciseness of the generated questions. Step 1, which focuses on incorporating domain knowledge, contributes more to generation effectiveness. Step 2 improves brevity by guiding the model to generate focused questions. Together, they enable the generation of sub-questions that are both informative and efficient for downstream QA tasks. A case study illustrating the model's generation evolution is presented in Appendix~\ref{sec:appendix_case_study}.

Importantly, removing Step 1 results in performance similar to no fine-tuning, emphasizing that domain knowledge is essential for EQD's effectiveness. Step 2 alone cannot optimize decomposition without the foundation built in step 1.

These findings highlight the necessity of our two-step strategy. It enables the EQD model to generate sub-questions that are both informative and efficient for downstream QA tasks.

\subsection{Efficiency Analysis}
\label{sec:efficiency_analysis}

\begin{figure}[t]
\vspace{-0.0cm}
\centering
\begin{center}
   \includegraphics[width=1.0\linewidth]{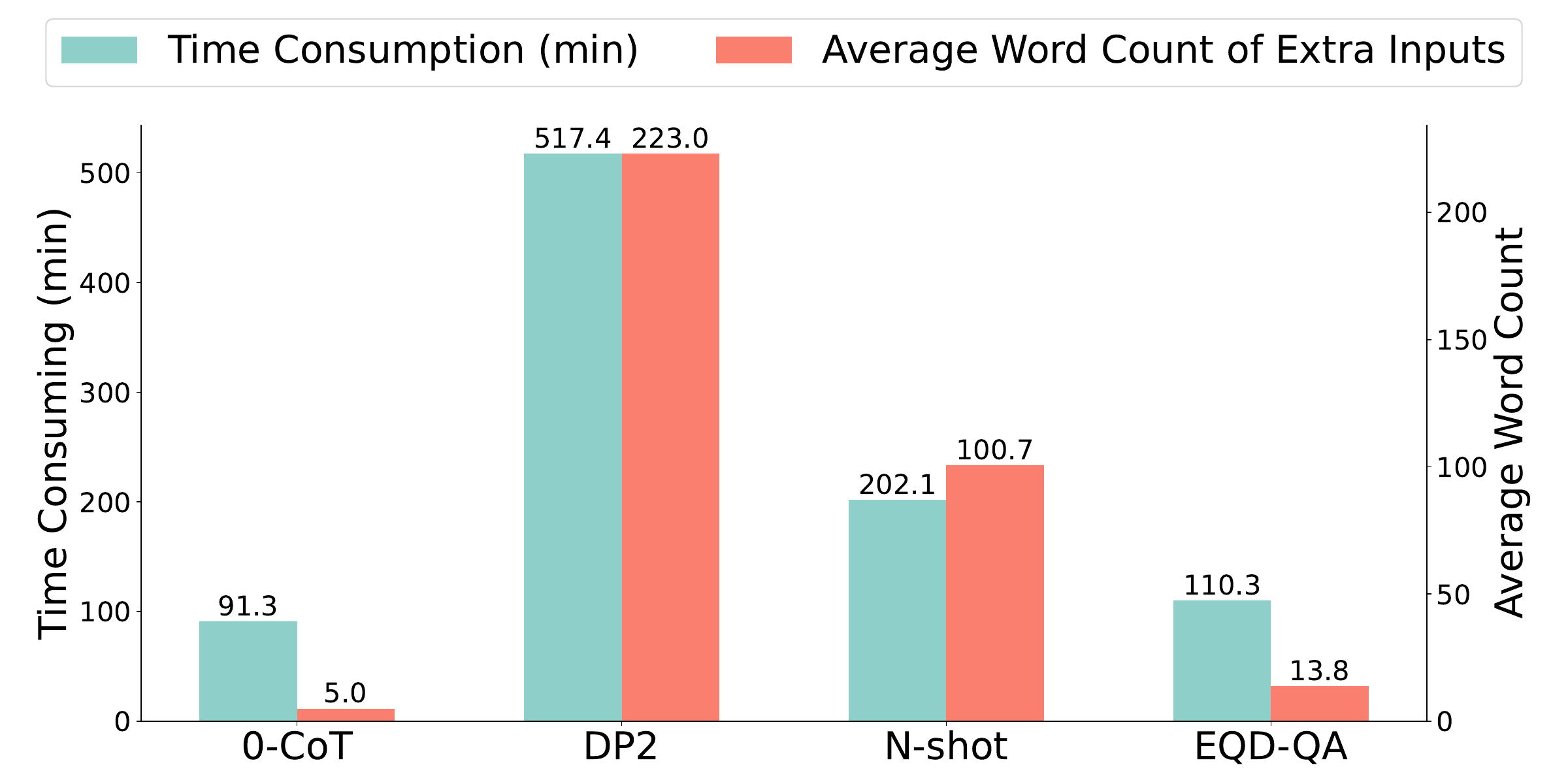}
    \vspace{-0.8cm}
   \captionof{figure}{Comparison of inference time consumption and input length across methods. Blue bars represent time consuming (left y-axis), while red bars indicate the average word count of extra inputs (right y-axis).}
\label{fig:efficiency}
\end{center}
\vspace{-0.6cm}
\end{figure}

Figure~\ref{fig:efficiency} presents a comparison of inference time consumption and additional input length across different methods on the test split of the FinQA dataset. We compare our EQD method with three baseline approaches, 0-CoT, DP2, and N-shot, which are the fastest among similar methods.

Our EQD-QA shows significantly lower inference time and shorter extra inputs compared to N-shot and DP2 (using GPT-4o as the QD model). Its inference time and extra prompt length are only marginally higher than 0-CoT, which simply adds reasoning prompts without domain knowledge. 

Further analysis shows GPT-4o generates an average of 7.3 supporting questions per case, while EQD generates just 1.2. As illustrated in Figure~\ref{fig:overview}, GPT-4o tends to produce overly detailed decomposed questions that may hinder reasoning. When considered with the previous performance comparisons, these results indicate that our EQD model generates more concise yet effective sub-questions for domain quantitative reasoning. The findings suggest that a single critical supporting question can be more beneficial for domain quantitative reasoning than multiple detailed reasoning steps.

\section{Conclusions}
\label{sec:conclusions}
This paper introduces a novel two-step fine-tuning method, including domain fine-tuning and QA expert alignment, to develop an Expert Question Decomposition (EQD) model. Our EQD model demonstrates significant effectiveness and efficiency in supporting domain quantitative reasoning. Experimental results across four financial datasets show consistent improvements in QA performance across various LLMs, maintaining computational efficiency comparable to zero-shot prompting methods. Furthermore, our analysis reveals that a single critical supporting question is more beneficial to the domain QA process than detailed step-by-step guidance, providing novel insights into LLMs' reasoning capabilities in specialized domains.

\section*{Limitations}
While our EQD model presents a novel approach to enhance LLMs' domain quantitative reasoning capabilities, two primary limitations should be acknowledged, due to resource constraints and data availability:

First, our baseline comparison includes only FinMA as a representative domain fine-tuned LLM, due to limited accessibility of similar models. For instance, FinLLaMA~\cite{xie2024open}, a recent financial domain fine-tuned LLM, requires author authorization for access, but inactive response to access requests has restricted its availability. However, this limitation does not impact our comparative analysis, as the reported performance of major financial LLMs (e.g., BloomBergGPT; \citealt{wu2023bloomberggpt}, and InvestLM; \citealt{yang2023investlm}) on FinQA and TAT-QA datasets falls considerably below our results using Claude3.5-sonnet. As demonstrated through FinMA, these domain fine-tuned models are inherently constrained by their base model capabilities and often lag behind state-of-the-art LLMs. In contrast, our EQD model's flexible integration with advanced LLMs offers advantages without such constraints.

Second, our evaluation of the EQD-QA method is evaluated only on the financial domain, a representative field for quantitative reasoning. The implementation of our method requires domain-specific question decomposition datasets for fine-tuning. At present, ConvFinQA~\cite{chen2022convfinqa} is the only publicly available dataset that meets our requirements. Although this limited dataset availability constrains broader evaluation across domains, our method's design and underlying principles are domain-agnostic, not specifically tied to financial knowledge. Moreover, since ConvFinQA contains only 3,037 training samples, datasets of this scale are feasible for companies to annotate based on their practical requirements. Therefore, our method has potential for extension to other domains.

\section*{Acknowledgments}
This work was supported by the UKRI Centre for Doctoral Training (CDT) in Natural Language Processing through UKRI grant EP/S022481/1.  We also acknowledge the support of the Centre for Investing Innovation at the University of Edinburgh.

\bibliography{main}

\appendix

\section{More implementation details}
\label{sec:appendix_implementation}

\subsection{Fine-tuning Settings}
We added a LoRA adapter with a rank of 8 and a LoRA alpha of 16. The fine-tuning process targeted eight parameter matrices: ``q\_proj'', ``k\_proj'', ``v\_proj'', ``o\_proj'', ``gate\_proj'', ``up\_proj'', ``down\_proj'', and ``lm\_head''. The adapter includes 22 million parameters, representing 0.27\% of the original model's parameters. 

We used a batch size of 32 for step 1 and 8 for step 2. The learning rate was set to 1e-5, with a warm-up of 5 steps. The maximum number of training iterations was set to 1000 for step 1 and 500 for step 2. Model checkpoints were selected based on performance on the FinQA validation set: iteration 400 for step 1 and iteration 200 for step 2.

During step 2, the average length of generated responses dropped from about 200 tokens at initialization to approximately 20 tokens within the first 50 iterations. It remained stable at around 20 tokens until iteration 500. The selected checkpoint in step 2 did not complete a full epoch over the FinQA training set. This is reasonable as the goal of step 2 is not to exhaustively learn all QA pairs, but to capture the reasoning challenges faced by the QA model. Although FinQA contains numerous QA pairs, many share similar reasoning structures. Step 2 prioritizes learning he difficulty of the reasoning process rather than memorizing knowledge from specific QA instances.

\subsection{Device and Training Time}
We used an A100 GPU for both training and testing. Step 1 fine-tuning required approximately 2.5 hours, and the step 2 required around 4 hours. Both the device and time requirements are significantly less than those for fine-tuning a domain-specific LLM.

\subsection{API Cost}
For all closed-source LLMs, including GPT-3.5-turbo, GPT-4o, o3-mini, and Claude3.5-sonnet, we used API to run experiments. Converting the two summary datasets, ECTSum and EDTSum, to QA datasets cost approximately \$2. Using these LLMs as QA models cost around \$200, primarily covered by GPT-4o and Claude3.5-sonnet. Using them as QD models cost approximately \$10.

\subsection{Evaluation Details}
Following prior work on FinQA dataset~\cite{chen2021finqa, singh2024finqapt}, we evaluate QA performance using the exact match accuracy (EmAcc) metric. Since most annotated answers are numerical, this evaluation process includes extracting answer strings using regular expression patterns, converting value representations to float numbers, matching digits between answers and ground truth, and performing comparisons.

Our reported results may differ from those in other papers due to differences in value extraction and evaluation implementations. Such inconsistencies are evident across multiple prior works, including InvestLM~\cite{yang2023investlm}, PIXIU~\cite{xie2023pixiu}, and FinQAPT~\cite{singh2024finqapt}, which report divergent FinQA scores for the same models, such as GPT-3.5. These discrepancies arise from the absence of publicly released evaluation code, leading to differences in implementation details. To ensure fairness and consistency, we evaluated all models using our own implementation. Our evaluation code has been released to promote standardized benchmarking practices.

\section{Baseline Methods}
\label{sec:appendix_baseline}
We implement several prompting methods as baselines for comparison: zero-shot Chain-of-Thought (0-CoT)~\cite{wei2022chain}, decomposed prompting (DP)~\cite{khot2022decomposed}, question decomposition CoT (QD-CoT)~\cite{zhou2022least}, retrieval CoT (R-CoT)~\cite{trivedi2022interleaving}, and few-shot in-context learning (N-shot)~\cite{li2023few}. The details of each method and their implementation are as follows:

\noindent\textbf{0-CoT}. Zero-shot CoT is a widely used prompting strategy to enhance LLMs' reasoning capabilities. It simply appends the phrase ``\emph{Let's think step by step.}'' to the end of a question, encouraging the model to provide intermediate reasoning steps before answering.

\noindent\textbf{DP}. Decomposed Prompting improves QA by breaking a complex question into sub-questions, which are then answered sequentially. This is a basic way of decomposing questions to improve QA performance, without domain adaptation and QA-specific optimization. It often produces overly detailed decompositions that may negatively impact answer accuracy. We implement this method using both GPT-3.5-turbo (DP1) and GPT-4o (DP2) to ensure a fair and comprehensive comparison. 

\noindent\textbf{QD-CoT}. This method combines few-shot learning with question decomposition. We implement it by selecting decomposition examples from the ConvFinQA training set and including them in the prompt to guide the model’s decomposition process.

\noindent\textbf{R-CoT}. Retrieval-based Chain-of-Thought enhances the QA process by incorporating retrieved external knowledge. We use the ConvFinQA dataset as the retrieval corpus and include relevant knowledge to support the model's reasoning.

\noindent\textbf{N-shot}. Few-shot in-context learning enhances reasoning through adding examples to the prompt. Following recent adaptations for financial data~\cite{singh2024finqapt}, we implement this method using examples from the training split of FinQA dataset. Our implementation uses sentence embeddings generated by OpenAI-Ada-002\footnote{\url{https://openai.com/blog/new-and-improved-embedding-model}} to identify similar questions from the training set. The annotated reasoning program steps from these similar questions are then incorporated as examples, guiding LLMs to generate analogous reasoning chains for new queries. We report the best results across 1-shot, 3-shot, and 5-shot settings.

\section{Prompt Design}
\label{sec:appendix_prompt}

We use two main types of prompts in our experiments: (1) prompts for QA and (2) prompts for question decomposition. The QA prompts guide the LLMs during answer generation. The decomposition prompts are used both for instruction tuning in the first fine-tuning step and for generating supporting questions during EQD inference.

\noindent\textbf{Question Answering Prompts}. The following base prompt is used to guide LLMs in the QA task:
\begin{spverbatim}
You are a financial expert capable of analyzing and answering financial questions based on the given context. Focus on extracting relevant numerical data, simplifying information, and providing concise answers.
\end{spverbatim}
\vspace{0.2cm}

We slightly adapt this prompt depending on the dataset to ensure consistent output formatting. For example, since the FinQA dataset only contains numerical answers or binary responses (yes/no), we add the following constraint:
\begin{spverbatim}
The final answer must include only a number (rounded to 5 decimal places), the word 'yes', or the word 'no', without any additional explanation or commentary.
\end{spverbatim}
\vspace{0.2cm}

\noindent\textbf{Question Decomposition Prompt}.
The following prompt is used for generating decomposed questions, both during model training and inference:
\begin{spverbatim}
You are a financial expert capable of analyzing financial questions. Break down this financial question into simpler sub-questions.
\end{spverbatim}

\section{Dataset Details}
\label{sec:appendix_dataset}
We conduct testing on four distinct financial datasets: FinQA, TAT-QA~\cite{zhu2021tat}, ECTQA, and EDTQA. These datasets cover both unstructured and tabular financial content, with varying lengths and source formats. Table~\ref{tab:datasets} summarizes the statistics of these test sets. 

\begin{table}[h]
\small
\vspace{0.15cm}
\begin{center}
\begin{tabular}[b]{l l r r}
\toprule
Dataset & Resource & Size & Avg. Words \\
\cmidrule{1-4}
FinQA  & Earning Reports     & 1147 & 700 \\
TAT-QA & Financial Reports   & 1663 & 220 \\
ECTQA  & Earning Transcripts & 1816 & 2715 \\
EDTQA  & Financial News      & 1662 & 714 \\
\bottomrule
\end{tabular}
\vspace{-0.2cm}
\captionof{table}{Statistics of QA test sets.}
\label{tab:datasets}
\end{center}
\vspace{-0.4cm}
\end{table}

ECTQA and EDTQA are derived from ECTSum~\cite{mukherjee2022ectsum} and EDTSum~\cite{xie2024finnlp}, which were originally summarization datasets. We convert them into QA datasets using GPT-4o, which generates questions based on the annotated summarization points. The goal of this conversion is to create QA pairs that focus on numerical reasoning. We used the following prompt to generate questions from individual summary sentences:
\begin{spverbatim}
Focus on the numerical value in the following sentence and convert the sentence into a question. The question should be answerable by a single number only. Provide your output in the following format: \nQuestion: []\nAnswer: []
\end{spverbatim}
\vspace{0.2cm}

To verify the quality of the generated QA pairs, we manually reviewed 400 examples from ECTQA and 200 from EDTQA. Six QA pairs were judged ambiguous or unclear for ECTQA and only two for EDTQA. The error rates of these two generated datasets are both under 2\%. Since each question is generated from a complete sentence, manual verification was straightforward and confirmed the effectiveness and correctness of the generated QA pairs.

\section{Fine-tuning Strategy}
\label{sec:appendix_ft_strategy}

\begin{table}[b]
\vspace{-0.2cm}
\begin{center}
\begin{tabular}[b]{l r}
\toprule
Strategy & EmAcc \\
\cmidrule{1-2}
Direct Fine-tuning & 42.5 \\
LoRA Continue & 54.0 \\
LoRA Extra & 44.8 \\
\bottomrule
\end{tabular}
\vspace{-0.2cm}
\captionof{table}{Comparison of EQD fine-tuning strategies using Llama3.1-8B-Instruct as the QA model on the FinQA dataset.}
\label{tab:strategy}
\end{center}
\vspace{-0.3cm}
\end{table}

\begin{table*}[t]
\centering
\begin{tabular}{p{0.9\textwidth}}
\toprule
\rowcolor{gray!10}\textcolor{black}{\textbf{Original Question}} \\
What is the growth rate in the balance of standby letters of credit from 2006 to 2007? \\
\rowcolor{gray!10}\textcolor{black}{\textbf{Decomposed Questions from the Base Llama3.1-8B-Instruct Model}} \\
To find the growth rate in the balance of standby letters of credit from 2006 to 2007, we need to break down the problem into simpler sub-questions:

1.  What is the balance of standby letters of credit in 2006?

2.  What is the balance of standby letters of credit in 2007?

3.  How do we calculate the growth rate?

Let's assume we have the required data. We'll use the following formula to calculate the growth rate:

Growth Rate = ((New Balance - Old Balance) / Old Balance) * 100

We'll need to know the balances for 2006 and 2007 to calculate the growth rate.

Sub-question 1: What is the balance of standby letters of credit in 2006?

- We need to know the value of the standby letters of credit at the end of 2006.

Sub-question 2: What is the balance of standby letters of credit in 2007?

- We need to know the value of the standby letters of credit at the end of 2007. \\
\rowcolor{gray!10}\textcolor{black}{\textbf{Decomposed Questions from the QD Model after Step 1 Fine-tuning}} \\
What was the change in the balance of standby letters of credit from 2006 to 2007?

And what was that balance in 2006?

How much, then, does that change represent in relation to this 2006 balance?\\

\rowcolor{gray!10}\textcolor{black}
{\textbf{Decomposed Questions from the EQD Model after Two-step Fine-tuning}} \\
What is the balance of standby letters of credit in 2006 and 2007?\\
\bottomrule
\end{tabular}
\vspace{-0.2cm}
\caption{The evolution of decomposed questions generated by the models after different fine-tuning steps.}
\label{tab:case_study}
\vspace{-0.2cm}
\end{table*}

For the QA expert alignment (second fine-tuning step), we implement three distinct strategies:

\begin{enumeratesquish}{0em}{0.02em}
    \item[1] ``Direct Fine-tuning'': This strategy involves fine-tuning a new LoRA adapter in step 2, independent of the LoRA adapter from step 1. During the second adapter's fine-tuning, the first adapter is removed. The final EQD model combines the base model with both LoRA adapters simultaneously.
    \item[2] ``LoRA Continue'': This method continuously fine-tunes the LoRA adapter from step 1.
    \item[3] ``LoRA Extra'': This method fine-tunes a new LoRA adapter in step 2 while keeping the LoRA adapter from step 1 locked and active. The final EQD model integrates the base model with both LoRA adapters.
\end{enumeratesquish}

We evaluate these fine-tuning strategies on the FinQA dataset using Llama3.1 as the QA model, with results presented in Table~\ref{tab:strategy}. The ``LoRA Continue'' strategy demonstrates superior performance compared to other methods. Based on these findings, we adopt ``LoRA Continue'' as our primary fine-tuning strategy for the EQD model.

\section{Results of EQD Models Fine-tuned from Different Base LLMs}
\label{sec:appendix_other_base}

We trained our EQD model using three additional LLMs of varying sizes and architectures: Llama3.2-1B-Instruct, Llama3.2-3B-Instruct, and DeepSeek-R1-Distill-Qwen-7B. All models were evaluated with Llama3.1-8B-Instruct as the QA model on the FinQA dataset. The results are summarized in Table \ref{tab:other_EQD}.

\begin{table}[b]
\vspace{-0.2cm}
\begin{center}
\begin{tabular}[b]{l r}
\toprule
EQD base model & EmAcc \\
\cmidrule{1-2}
Llama3.2-1B-Instruct & 52.3 \\
Llama3.2-3B-Instruct & 53.9 \\
DeepSeek-R1-Distill-Qwen-7B & 54.3 \\
\bottomrule
\end{tabular}
\vspace{-0.2cm}
\captionof{table}{Comparison of EQD models fine-tuned from different base LLMs.}
\label{tab:other_EQD}
\end{center}
\vspace{-0.3cm}
\end{table}

Compared with the results in Table \ref{tab:llm_resuls}, these findings show that although model size affects EQD performance, even smaller models remain highly effective. All EQD-enhanced results on FinQA substantially outperform direct QA without EQD. The 1B model performs slightly worse than the larger models, while the 3B model achieves performance comparable to the 8B model, demonstrating good scalability. Model architecture also has minimal impact on EQD effectiveness, as the DeepSeek-based EQD model performs similarly to the Llama-based model of comparable size. Overall, these results confirm that our method is robust and effective even when applied to smaller LLMs.

\section{Case Study}
\label{sec:appendix_case_study}

\subsection{Evolution of Generated Decomposed Questions}

We fine-tune a base LLM, Llama3.1-8B-Instruct, by our two-step fine-tuning to an expert question decomposition model. The case study in Table~\ref{tab:case_study} illustrates the model's progressive transformation into an expert decomposition system.

\begin{table*}[t]
\centering
\begin{tabular}{p{0.9\textwidth}}
\toprule
\rowcolor{gray!10}\textcolor{black}{\textbf{Original Question}} \\
What is the growth rate in the balance of standby letters of credit from 2006 to 2007? \\

\rowcolor{gray!10}\textcolor{black}
{\textbf{Decomposed Questions from the EQD Model after Two-step Fine-tuning}} \\
What is the balance of standby letters of credit in 2006 and 2007?\\

\rowcolor{gray!10}\textcolor{black}
{\textbf{Answers to the Decomposed Questions}} \\
According to the given context, the balance of standby letters of credit in 2006 and 2007 are as follows: 2006: \$4,926 million 2007, \$4,711 million \\

\rowcolor{gray!10}\textcolor{black}
{\textbf{Answers to the Original Question}} \\
To find the growth rate, we need to calculate the difference in the balance of standby letters of credit between 2007 and 2006, and then divide it by the balance in 2006.

Balance in 2006: \$4,926 million
Balance in 2007: \$4,711 million*

Difference: \$4,926 million - \$4,711 million = -\$215 million (negative, indicating a decrease)

Growth rate: (\$215 million / \$4,926 million)  100\% $\approx$ -4.36\%

The balance of standby letters of credit decreased by approximately 4.36\% from 2006 to 2007.\\

\bottomrule
\end{tabular}
\vspace{-0.2cm}
\caption{Answers generated during the QA process using our EQD model, including both the answers to the decomposed sub-questions and the final answer to the original question.}
\label{tab:case_study2}
\vspace{-0.2cm}
\end{table*}

The base Llama3.1-8B-Instruct model initially generates overly detailed reasoning guidance, similar to other LLMs like GPT models shown in Figure~\ref{fig:overview}. After Step 1 (domain fine-tuning), the decomposed questions become more concise and professional, focusing on key steps and domain-specific terms of the original question. Upon completing the full two-step fine-tuning process, the generated questions become highly targeted, with a single question addressing the core challenge in answering the original query.

\subsection{Answer Example}
Building on the practical case of demonstrating the evolution of decomposed questions, we present the answering process for the final question in Table~\ref{tab:case_study2}.

\end{document}